\RequirePackage{placeins}
\RequirePackage{xurl}
\PassOptionsToPackage{table}{xcolor}
\RequirePackage{xcolor}
\documentclass[journal,twoside,web,notranslateabstract]{ieeecolor}
\usepackage{generic}
\usepackage{cite}
\usepackage{amsmath,amssymb,amsfonts}
\usepackage{algorithm,algorithmic}
\usepackage{multirow}
\usepackage{graphicx}
\usepackage{hyperref}
\hypersetup{hidelinks}
\usepackage{textcomp}
\usepackage{booktabs,float,placeins,array,etoolbox}
\providecommand{\refname}{References}
\usepackage[normalem]{ulem}
\usepackage{xurl}
\usepackage{tikz}
\usetikzlibrary{arrows.meta,positioning,shapes.geometric}
\DeclareUnicodeCharacter{00D7}{\ensuremath{\times}}
\DeclareUnicodeCharacter{00B1}{\ensuremath{\pm}}
\graphicspath{{figures/}{./}}

\newcolumntype{L}[1]{>{\raggedright\arraybackslash}p{#1}}
\def\BibTeX{{\rm B\kern-.05em{\sc i\kern-.025em b}\kern-.08em
 T\kern-.1667em\lower.7ex\hbox{E}\kern-.125emX}}
\makeatletter
\patchcmd{\thebibliography}{\itemsep 0pt plus pt}{\itemsep 0pt plus 0.1pt}{}{}
\patchcmd{\thebibliography}{\section*{\color{black}}}{\section*{References}}{}{}
\long\def\@makecaption#1#2{%
  \begingroup\normalfont\sffamily\fontsize{8}{10}\selectfont
  \ifx\@captype\@IEEEtablestring
    \@IEEEtablecaptionsabspace
    {\centering\scshape #1\par\vskip2pt #2\par}%
    \@IEEEtablecaptionsbespace
  \else
    \@IEEEfigurecaptionsepspace
    {\raggedright #1.\enspace #2\par}%
  \fi\endgroup}

\def\ps@titlepagestyle{%
  \def\@oddfoot{}\def\@evenfoot{}%
  \def\@oddhead{\vbox{\hsize\textwidth
    \hbox to \textwidth{\includegraphics[height=14pt]{\logoname.eps}\hfil
      \parbox[b]{.85\textwidth}{\scriptsize\sffamily\raggedleft\leftmark\quad\thepage}}
    \vskip3pt{\color{subsectioncolor}\hrule height1pt}}}%
  \let\@evenhead\@oddhead}
\makeatother
\begin{document}
\title{MMDA-BCST: Micro-Scale Searching Multi-Agent Data Augmentation for Biologically Consistent Spatial Transcriptomics Prediction}
% The original manuscript does not supply author names or affiliations.
\author{Fangqing Liu, \IEEEmembership{member, IEEE}, Lingshuang Tang, Yingrui Chen, Ziyu Ning, Chenyi Xie, Qingling Zhang, Baoxiong Zhuang, and Han Huang*, \IEEEmembership{Senior Member, IEEE}
\thanks{This work is supported by National Natural Science Foundation of China (62576116, 92570116, 82302299).
F. Liu is with the School of Management, Guangdong University of Technology, Guangzhou, 510520 China, Email: peerfog@gdut.edu.cn.
%F. Liu is with the School of Management, Guangdong University of Technology, Guangzhou 510520, China, and also with the Key Laboratory of Symbolic Computation and Knowledge Engineering of Ministry of Education, Jilin University, Changchun 130012, China, Email: peerfog@gdut.edu.cn.
L. Tang is with the Zhongshan School of Medicine, Sun Yat‑sen University, Guangzhou, 510080 China, Email: lingshuangtang@outlook.com.
Y. Chen is with the School of Computer Science and Technology, Guangdong University of Technology, Guangzhou, 510006 China, Email: chenyingrui@mails.gdut.edu.cn.
Z. Ning, C. Xie, and Q. Zhang are with the Department of Pathology, Guangdong Provincial People's Hospital (Guangdong Academy of Medical Sciences), Southern Medical University, Guangzhou, 510080 China, Email: ningziyu@gdph.org.cn, xiechenyi@gdph.org.cn, and zhangqingling@gdph.org.cn.
B. Zhuang is with the Department of General Surgery \& Guangdong Provincial Key Laboratory of Precision Medicine for Gastrointestinal Tumor, Nanfang Hospital, The First School of Clinical Medicine, Southern Medical University, Guangzhou, Guangdong, 510515 China. Email: 18820095121@163.com
H. Huang is with the School of Software Engineering, Sun Yat-sen University, Zhuhai, 519082 China. Email: huangh985@mail.sysu.edu.cn}}
\maketitle

\begin{abstract}
H\&E-to-gene-expression prediction for spatial transcriptomics provides a promising approach to characterizing the tissue micro-environment. However, this task is challenged by scarce paired samples, and existing augmentation methods may introduce biologically inconsistent samples that disrupt tissue structure, alter staining, or distort spatial gene-expression relationships. To alleviate data scarcity while preserving biological consistency, we propose MMDA-BCST, a multi-agent data augmentation framework via micro-scale searching for spatial transcriptomics prediction. In MMDA-BCST, augmentation-strategy discovery is formulated as a constrained sequential decision process. Each candidate is generated, evaluated, and checked for biological consistency before acceptance. Agents coordinated through shared memory handle candidate generation, predictive evaluation, and biological-consistency verification. They use positive and negative evidence to avoid redundant probing. We evaluate MMDA-BCST on three spatial transcriptomics datasets with four downstream models and ten augmentation categories. MMDA-BCST improves downstream prediction under scarce paired samples and reduces the risk of biologically inconsistent augmented samples. Ablation results show that gains mainly arise from micro-scale searching rather than from increasing the number of agents. Coordination narrows the search to promising candidates and reduces redundant computation.
\end{abstract}

\begin{IEEEkeywords}
Spatial transcriptomics, histopathology image, data augmentation, multi-agent systems, micro-scale searching
\end{IEEEkeywords}

\section{Introduction}\label{sec:introduction}

Spatial transcriptomics measures high-dimensional gene expression in situ while preserving spatial coordinates, enabling the study of tumor microenvironments, tissue architecture, and cell states \cite{stahl2016spatial,hu2021spagcn,bergenstrahle2022super}. As H\&E-stained images are routine in pathology and central to digital pathology representation learning \cite{xu2024gigapath}, predicting spatial gene expression from these images could complement costly molecular assays. Two lines of work are most relevant. The first line, including HE2Gene \cite{he2020integrating}, Hist2ST \cite{zeng2022hist2st}, iStar \cite{zhang2024istar}, HisHRST \cite{shi2025hishrst}, and FineST \cite{li2026finest}, extends joint image-expression modeling from multi-task and spatially constrained prediction to super-resolution, hierarchical representation, and fine-grained expression maps. The second line, including GHIST \cite{fu2025ghist}, NH2ST \cite{liu2025nh2st}, TRIPLEX \cite{chen2024triplex}, and HESpotEx \cite{yin2026hespotex}, focuses on single-cell prediction, neighborhood learning, multi-resolution fusion, and multi-tissue validation. Collectively, these studies demonstrate consistent progress in spatial modeling, multi-scale representation, and cross-sample generalization.

Despite these advances, limited paired samples and tissue heterogeneity, together with the complex relationship between morphology and high-dimensional expression, continue to constrain performance and generalization \cite{wang2025benchmarking}. Data augmentation is widely used to alleviate the scarcity of paired samples \cite{shorten2019survey}. Representative methods include AutoAugment \cite{cubuk2019autoaugment}, RandAugment \cite{cubuk2020randaugment}, and TrivialAugment \cite{muller2021trivialaugment}. These methods automatically search for or simplify augmentation strategies originally designed for natural images. CutMix \cite{yun2019cutmix} and AugMix \cite{hendrycks2020augmix} expand training distributions through sample mixing or augmentation chains. RandStainNA \cite{shen2022randstainna}, CLASS-M \cite{zhang2025classm}, and ADA \cite{huang2025ada} improve staining robustness, structural consistency, or adaptability in pathology and medical imaging. However, when applied to expression prediction in spatial transcriptomics, these augmentation methods share a common limitation. They treat augmentation as a fixed, isolated preprocessing step and do not update augmentation decisions based on prediction feedback. These limitations point to three gaps. First, when paired samples are scarce, general operators and fixed recipes provide limited adaptation to sample characteristics, downstream models, and biological constraints. Second, prediction scores alone are insufficient to verify whether the tissue morphology, spatial neighborhoods, and staining patterns of augmented samples remain consistent with expression supervision \cite{tellez2019quantifying}. Third, failed trials and repair outcomes are rarely recorded or reused, potentially wasting training budgets on repeated trials. These gaps suggest that scarce-sample augmentation requires not only efficient search but also reliable evidence reuse and biological verification.

To address these gaps, we propose the following three hypotheses. H1: Role-separated multi-agent collaboration improves the reliability of feasible-strategy discovery compared with single-agent search or agent-free search. H2: Independent biological feedback identifies high-scoring false winners that violate tissue, staining, or spatial-expression constraints. H3: Shared memory reduces repeated trials and narrows the search space in later rounds by retaining positive, negative, and repair evidence, thereby improving search efficiency under scarce paired samples. These hypotheses motivate our contributions and guide the experiments reported later.

To investigate these hypotheses, we propose a multi-agent data augmentation framework driven by micro-scale searching for biologically consistent spatial transcriptomics prediction (MMDA-BCST). It coordinates three agents through a shared memory layer and focuses computation on informative candidates through staged evaluation and trial deduplication. Our contributions are summarized as follows.
\begin{itemize}
\item  {\textbf{A constrained sequential decision model for biologically consistent augmentation.} We formulate augmentation-strategy discovery as a sequential decision process under biological-consistency constraints. By decoupling predictive gain from biological feasibility, this model explicitly excludes high-scoring false winners.}
\item  {\textbf{A role-separated multi-agent algorithm driven by micro-scale searching.} We develop MMDA-BCST to solve this model. By restricting the search to local edits and separating agent roles, the algorithm avoids combinatorial explosion and confirmation bias, enabling efficient exploration and reliable evidence reuse.}
\item  {\textbf{Empirical evidence and domain impact for reliable augmentation.} Experiments show that MMDA-BCST improves downstream prediction under limited paired samples while reducing biological inconsistency risks. Under an identical budget, it accelerates search by 1.7\(\times\) to 10.0\(\times\) across datasets without sacrificing performance. This confirms that gains stem from closed-loop micro-scale searching rather than merely increasing the number of agents, offering a practical paradigm for biology-aware medical image augmentation.}
\end{itemize}

The remainder of this paper is organized as follows. Section~\ref{sec:background} formalizes the task and the augmentation-search problem. Section~\ref{sec:method} details the proposed framework and its complexity. Section~\ref{sec:experiments} presents research questions and evaluation results. Section~\ref{sec:conclusion} concludes and discusses future work.

\section{Background}\label{sec:background}

To delineate the scope of this study, this section formalizes the augmentation-search problem. Related prediction and augmentation methods are reviewed in a later subsection.

\subsection{Problem Formalization}\label{subsec:bg-formalization}

This study focuses on discovering training-time augmentation strategies for predicting spatial gene expression from H\&E images, rather than proposing a new expression-prediction network.

Given a spatial transcriptomics dataset $\mathcal{D}=\{(x_i,s_i,g_i)\}_{i=1}^{n}$, where $x_i$ is an H\&E patch, $s_i$ is the spatial coordinate, and $g_i$ is the gene expression vector, the dataset is partitioned into mutually disjoint training, validation, and locked test sets $\mathcal{D}_{\mathrm{tr}}$, $\mathcal{D}_{\mathrm{val}}$, and $\mathcal{D}_{\mathrm{te}}$, satisfying $\mathcal{D}=\mathcal{D}_{\mathrm{tr}}\cup\mathcal{D}_{\mathrm{val}}\cup\mathcal{D}_{\mathrm{te}}$. An augmentation strategy $\pi\in\Pi$ is represented as an ordered, variable-length operator graph in which each node records operator type, execution order, application probability, strength, and parameters. A candidate downstream predictor is denoted by $f_m$, where $m\in\mathcal{M}$ is the model identifier. The closed-loop state at round $t$ is defined as $S_t=(\mathcal{R},\mathcal{H}_t,b_t)$, where $\mathcal{R}$ is the predefined research setting, $\mathcal{H}_t$ is the structured evidence of completed trials, and $b_t$ is the remaining real training budget.

In MMDA-BCST, augmentation-strategy discovery is formulated as a sequential decision problem subject to biological-consistency constraints:
\begin{equation}
\footnotesize
a_t^*=\arg\max_{a\in\mathcal{A}(S_t)}\mathbb{E}[z_{t+1}\mid a,S_t],\quad \mathrm{s.t.}\quad \mathrm{Cost}(a)\leq b_t
\label{eq:sequential-decision}
\end{equation}
where $z_{t+1}$ indicates whether the candidate obtained after executing action $a_t$ achieves predictive gain while satisfying biological feasibility. Its formal definition is given in Eq.~\eqref{eq:feasibility}. $\mathcal{A}(S_t)$ is the set of admissible local strategy edits at the current state, including inserting, removing, or reordering operators, tuning the probability or strength, repairing paired gene-expression labels, enforcing biological consistency, or terminating the search. The action $a_t$ produces a candidate strategy $\pi_t$ together with its identity control $\pi_{\mathrm{ctrl},t}$. Both share the same $f_m$, data split, initialization, data order, and random seed. They differ only in the augmentation intervention. Actual execution yields structured evidence
\begin{equation}
\footnotesize
E_{t+1}=\operatorname{Execute}(f_m,\pi_t,\pi_{\mathrm{ctrl},t};\mathcal{D}_{\mathrm{tr}},\mathcal{D}_{\mathrm{val}})
\label{eq:evidence-execution}
\end{equation}
from which the paired predictive gain and biological-consistency vector are computed as
\begin{equation}
\footnotesize
\Delta Y_t=Y(\pi_t)-Y(\pi_{\mathrm{ctrl},t}),\qquad
\mathbf{q}_t=(q_t^{\mathrm{img}},q_t^{\mathrm{spa}},q_t^{\mathrm{exp}}).
\label{eq:evidence-components}
\end{equation}
Let $Y(\cdot)$ be the predefined validation metric, $\delta$ the minimum effective predictive gain, and $\boldsymbol{\tau}=(\tau_{\mathrm{img}},\tau_{\mathrm{spa}},\tau_{\mathrm{exp}})$ the threshold vector over the image-, spatial-, and expression-level biological checks. The candidate feasibility is then defined as
\begin{equation}
\footnotesize
z_t=\mathbb{I}(\Delta Y_t\geq\delta)\prod_{r\in\{\mathrm{img},\mathrm{spa},\mathrm{exp}\}}\mathbb{I}(q_t^r\geq\tau_r).
\label{eq:feasibility}
\end{equation}

When $z_t=1$, the candidate is both predictively beneficial and biologically feasible. When $\Delta Y_t\geq\delta$ but $z_t=0$, the candidate is defined as a false winner. After execution, $E_{t+1}$, $\Delta Y_t$, $\mathbf{q}_t$, $z_t$, and the failure reasons are written into $\mathcal{H}_{t+1}$ and, through shared memory, used to update the admissible action set of the next round. This formulation unifies data, strategy, budget, predictive gain, and biological constraints into a single sequential decision problem, without embedding unvalidated information- or coverage-based heuristics directly into the main objective.

\subsection{Related Works}\label{subsec:rw-gap}

Early studies established the feasibility of histology-based spatial expression prediction. HE2Gene \cite{he2020integrating} used multi-task learning and spatial constraints. Hist2ST \cite{zeng2022hist2st} combined convolutional and graph-neural modules for image and spatial modeling. iStar \cite{zhang2024istar} used hierarchical histological features for super-resolution inference. STMCL \cite{shi2025stmcl} fused images, expression, and locations via multimodal contrastive learning. Later works extended these settings. GHIST \cite{fu2025ghist} integrated cell morphology, type, neighborhood, and single-cell expression. NH2ST \cite{liu2025nh2st} modeled target-spot and neighborhood relations via dual-scale contrastive learning. TRIPLEX \cite{chen2024triplex} fused local and global multi-resolution features. However, none of these methods treats augmentation as a learnable component. Augmentation remains an external configuration or fixed preprocessing step. These studies confirm feasibility but focus on predictor architecture and spatial context. Since data sources, preprocessing, and resolution can affect benchmark conclusions \cite{wang2025benchmarking}, data augmentation warrants rigorous evaluation, especially with limited paired samples.

Image augmentation research offers mature strategies. AutoAugment \cite{cubuk2019autoaugment} searches policy spaces. RandAugment \cite{cubuk2020randaugment} and TrivialAugment \cite{muller2021trivialaugment} reduce tunable dimensions. CutMix \cite{yun2019cutmix} and AugMix \cite{hendrycks2020augmix} expand training distributions via region mixing and augmentation chains. For pathology, RandStainNA \cite{shen2022randstainna} models staining variation, and HED analysis \cite{tellez2019quantifying} characterizes staining differences. CLASS-M \cite{zhang2025classm} uses stain-separated MixUp, and ADA \cite{huang2025ada} adapts augmentation per sample. These methods provide useful operators, but their objectives do not target spatial expression supervision. Augmented samples are not verified for tissue morphology, spatial positions, or expression consistency.

Medical-image augmentation also emphasizes structure preservation, domain translation, and generalization with limited annotations. DDA-GAN \cite{chen2021dda} expands segmentation data via cross-modality generation. AdvChain \cite{chen2022advchain} applies photometric and geometric perturbations. Poisson blending \cite{wang2022poisson} creates local lesions, and MCFN \cite{ding2024mcfn} performs online augmentation for pathology classification. CellMix \cite{zhang2025cellmix} and positive-instance augmentation \cite{mi2025micl} further address pathology-specific scenarios. More recent advances explore implicit augmentation, semantic preservation, spectral intervention, deformation latent spaces, automatic augmentation, and test-time generation \cite{wan2025crida,ye2025fsdadg,liu2025spectrum,hossain2025mgaug,liu2026medaugment,ma2026ttga}. Generative approaches further improve out-of-distribution fairness, WSI-tile synthesis, large-scale tumor generation, and counterfactual-robust learning \cite{ktena2024generativefairness,carrilloperez2024syntheticwsi,wu2025freetumor,roschewitz2025counterfactual,wang2024dropshortcuts}. However, these methods target visual plausibility, accuracy, or generic robustness, not biological consistency with expression measurements.

In summary, spatial transcriptomics prediction models have shown the value of H\&E images, and augmentation research offers diverse operators. However, existing methods neither learn augmentation strategies under scarce paired samples nor verify biological consistency after augmentation, and they rarely reuse failure and repair evidence. These limitations call for a framework that learns strategies under scarce paired samples, verifies biological consistency, and reuses failure and repair evidence. We thus propose MMDA-BCST, integrating candidate generation, independent biological verification, and evidence reuse.

\section{Method}\label{sec:method}
\begin{figure*}[!t]
    \centering
    \includegraphics[width=15cm]{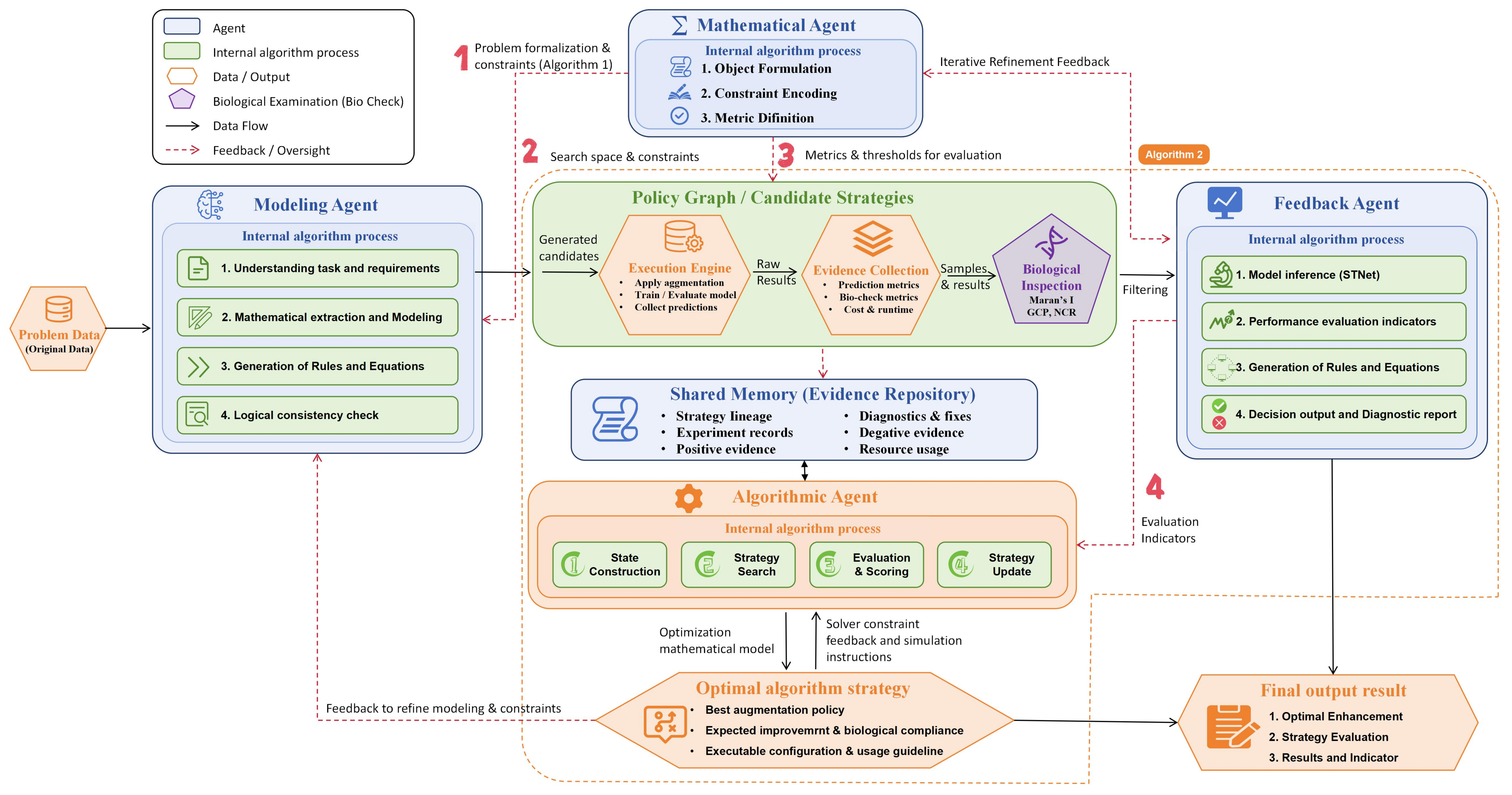}
    \caption{\textbf{Overview of the proposed MMDA-BCST framework.}}
    \label{Figure2}
\end{figure*}

As illustrated in Fig.~\ref{Figure2}, MMDA-BCST implements the micro-scale searching principle with three role-separated agents around a shared memory layer, namely modeling, algorithm, and feedback agents.

\subsection{Micro-scale Searching Principle and Search-Space Restriction}\label{subsec:micro-search-principle}

The augmentation-strategy space is jointly determined by operator type, order, probability, strength, model, and sample context. Enumerating all combinations directly would require many training runs that are either difficult to attribute or known to be risky. MMDA-BCST therefore first defines a search boundary to exclude out-of-scope combinations, then selects a local candidate subset from the remaining space. Each candidate is trained with its matched control, and it is accepted, repaired, or rejected by independent feedback based on predictive gain and biological consistency. In this paper, \emph{micro-scale searching} refers to a sequential search that, under a fixed data split, candidate model, operator scope, evaluation criterion, and biological thresholds, proposes only a few local edits around the current strategy and updates the next-round candidate scope based on real paired trials and the positive and negative evidence in shared memory \cite{huang2024correlation}. Micro-scale searching is not a new predictive model, nor does it replace training with surrogate metrics. Instead, it decides which candidates should consume the real training budget.

\begin{figure}[!t]
    \centering
    \includegraphics[width=\columnwidth]{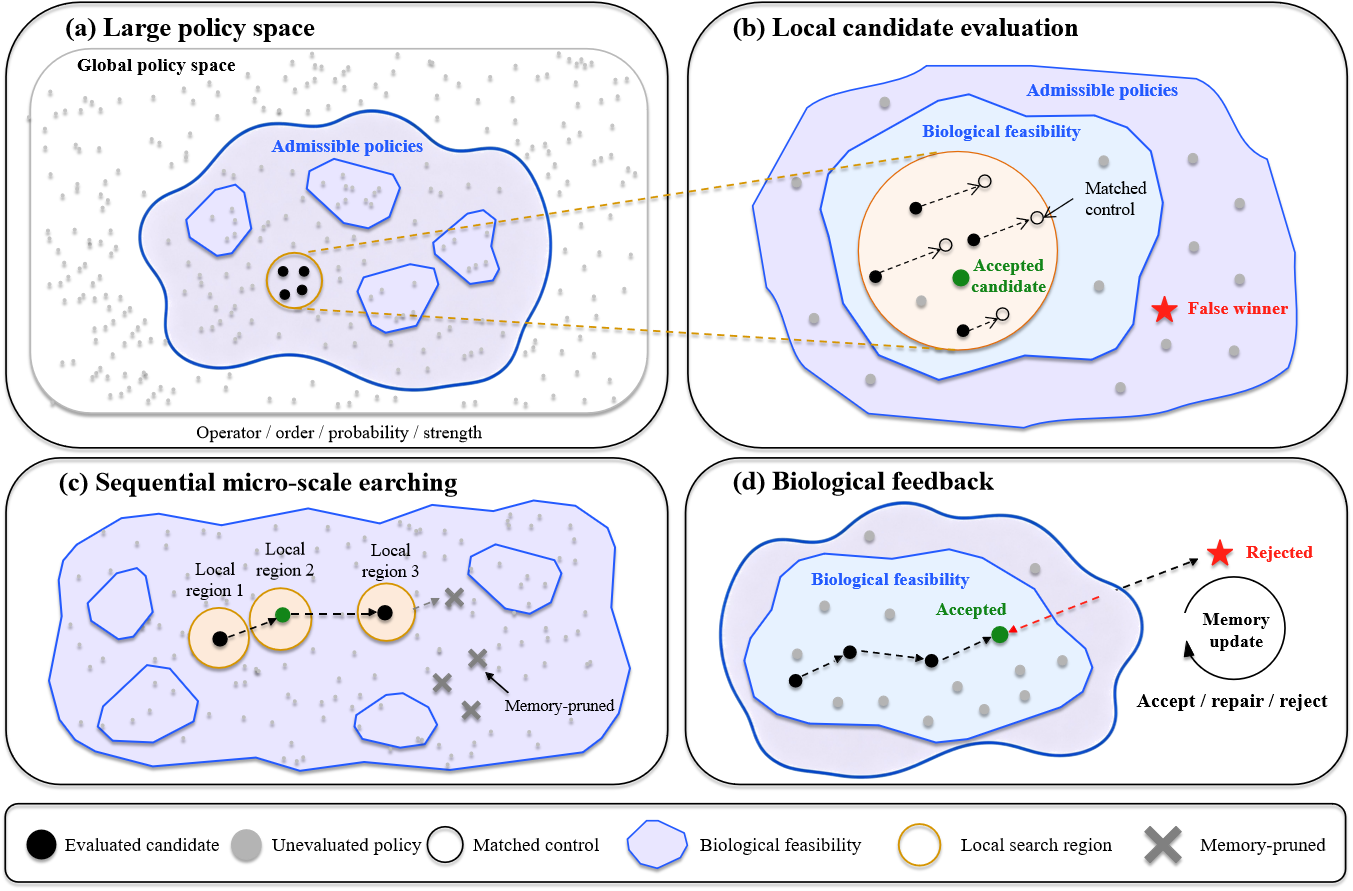}
    \caption{\textbf{Micro-scale searching in a constrained strategy space.}
\footnotesize(a) Candidate strategies vary by operator, order, probability, and strength. (b) Candidates are evaluated against no-augmentation controls within the biologically feasible region. (c) Historical failures iteratively refine the search scope. (d) Candidates are accepted, repaired, or rejected by biological feedback, and the search memory is updated.}
    \label{fig:micro-search-concept}
\end{figure}

Formally, let $\Omega_{\mathcal{R}}$ be the searchable strategy space, $\pi_{t-1}$ the strategy at round $t-1$, and $\mathcal{A}^{\mathrm{allow}}_t\subseteq\mathcal{A}(S_t)$ the allowed edit actions. The candidate pool entering training at round $t$ is then
\begin{equation}
\footnotesize
\mathcal{P}_t=\left\{\operatorname{Edit}(\pi_{t-1},a):a\in\mathcal{A}^{\mathrm{allow}}_t\right\}\subseteq\Omega_{\mathcal{R}},
\qquad |\mathcal{P}_t|\ll|\Omega_{\mathcal{R}}|.
\label{eq:local-candidate-pool}
\end{equation}

Local edits include insertion, deletion, reordering, probability adjustment, and strength refinement. As shown in Fig.~\ref{fig:micro-search-concept}(b)--(d), out-of-scope combinations are first excluded according to the research setting. In each round, only the candidate pool and its identity controls are trained, and verified evidence is used to update $\mathcal{A}^{\mathrm{allow}}_{t+1}$. This round-wise narrowing avoids equal-fidelity training over all combinations and concentrates high-fidelity training on candidates that remain both predictively beneficial and biologically feasible.

\subsection{Role-Separated Multi-Agent Coordination}\label{subsec:protocol}

\begin{figure}[!h]
\centering
\includegraphics[width=1\columnwidth]{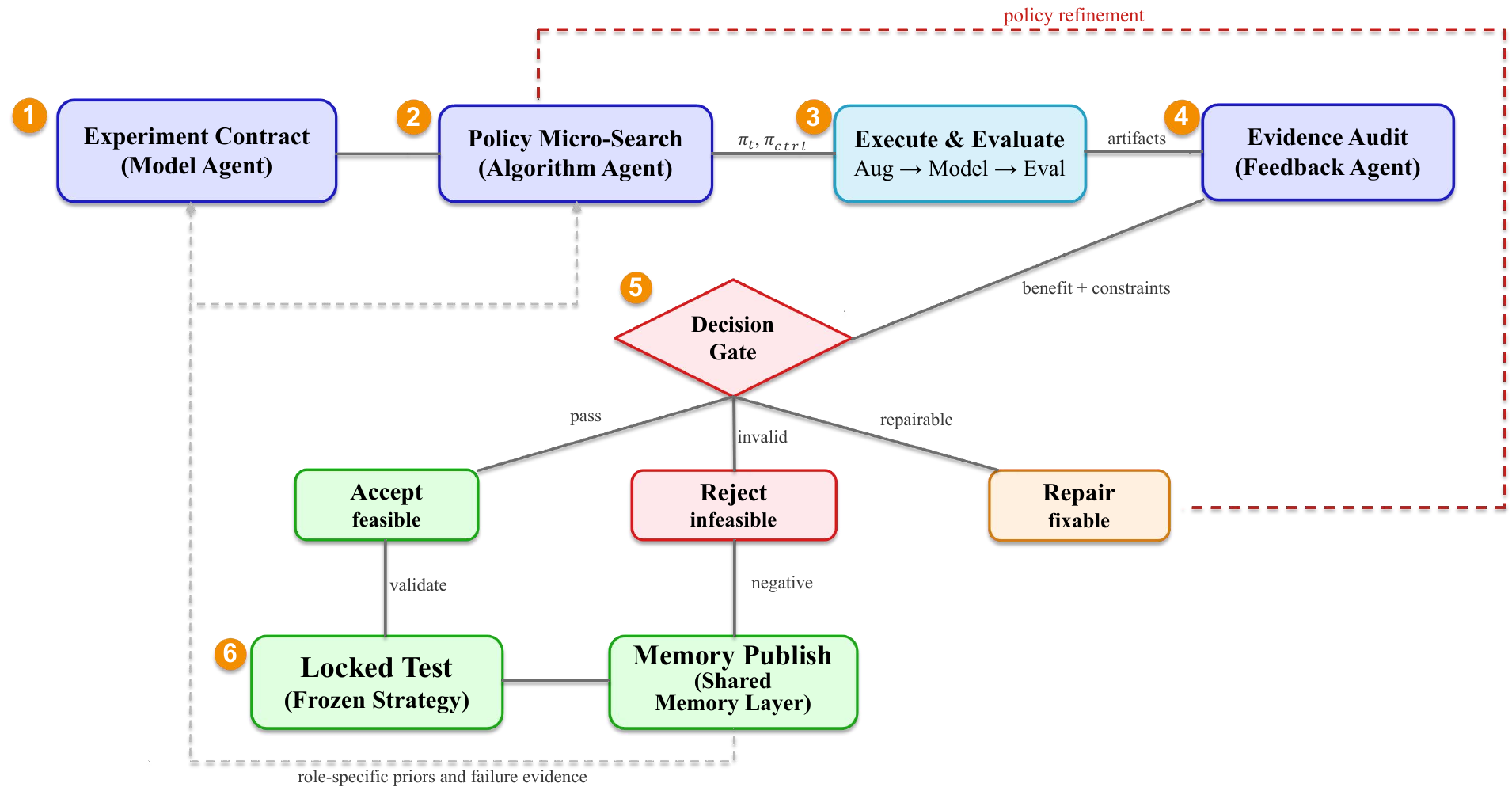}
\caption{\textbf{Closed-loop workflow of MMDA-BCST.}}
\label{fig:workflow_diagram}
\end{figure}

In this section, we specify who proposes candidates, verifies executions, and narrows the search space afterwards. A single module that simultaneously proposes strategies, performs computation, and certifies conclusions would structurally amplify confirmation bias. MMDA-BCST therefore decomposes the search kernel into four complementary components. These are a modeling agent, an algorithm agent, a feedback agent, and a shared memory layer. As shown in Fig.~\ref{fig:workflow_diagram}, in each round, the modeling agent supplies the frozen research setting, while shared memory retrieves role views of similar contexts. The algorithm agent then proposes candidates with matched controls, and the tool chain executes augmentation, training, and evaluation. Finally, the feedback agent reads only the frozen execution artifacts and issues an accept, repair, or reject decision. The evidence package is then written back to shared memory to update the action set of the next round. No agent may simultaneously modify the candidate, the execution result, and the final decision.

\subsubsection{Modeling Agent}\label{subsec:framework}

The modeling agent translates an open-ended research goal into a pre-declared research setting. It reads the data profile $\mathcal{D}_{\mathrm{profile}}$, task semantics $\mathcal{S}$, natural-language constraints $\mathcal{L}$, and context signature $\sigma$. Combining these inputs with a role-restricted modeling view $\mathcal{V}_{\text{model}}$ retrieved from shared memory, it determines the research setting $\mathcal{R}=(\mathcal{O},\mathcal{T}_{\text{AST}},\Omega_{\text{op}},\mathcal{M},\mathcal{B})$. The objective vector $\mathcal{O}$ weights the primary and auxiliary metrics. The constraint AST $\mathcal{T}_{\text{AST}}$ encodes spatial autocorrelation, staining drift, and tissue-region preservation into executable predicates. The candidate operator space $\Omega_{\text{op}}$ prunes the global augmentation library. The model domain $\mathcal{M}$ fixes the downstream architectures. The budget configuration $\mathcal{B}$ specifies the candidate pool size, search rounds, fidelity-promotion rules, and locked-test conditions. Algorithm~\ref{alg:contract_generation} generates $\mathcal{R}$, which enters the search only after passing deterministic validation by the feedback agent. The modeling agent cannot execute training, modify numerical results, or judge candidate validity. It may only narrow the scope or rewrite constraints when a conflict is detected.

\begin{algorithm}[t]
\caption{\small Biological Constraint Formalization and Search Space Initialization}
\label{alg:contract_generation}
\sloppy
\scriptsize
\begin{algorithmic}[1]
\REQUIRE Data profile $\mathcal{P}$, task semantics $\mathcal{S}$, natural-language constraints $\mathcal{L}$, context signature $\sigma$.
\ENSURE Structured experiment contract $C$.
\STATE Retrieve validated priors $\mathcal{V}_{\text{model}}$ from shared memory based on $\sigma$.
\STATE Extract previously validated operators $\Omega_{\text{prior}}$ from $\mathcal{V}_{\text{model}}$.
\STATE Initialize constraint AST $\mathcal{T}_{\text{AST}} \leftarrow \emptyset$.
\FOR{$\forall \text{ constraint } c \in \mathcal{L}$}
    \STATE Parse $c$ into an AST node and add to $\mathcal{T}_{\text{AST}}$.
\ENDFOR
\STATE Formulate objective weights $\mathcal{O}$ and define model domain $\mathcal{M}$ from $\mathcal{S}$ and $\mathcal{P}$.
\STATE Prune global operator library to $\Omega_{\text{op}}$ using $\Omega_{\text{prior}}$ and $\mathcal{T}_{\text{AST}}$.
\STATE Assemble contract $C \leftarrow (\mathcal{O}, \mathcal{T}_{\text{AST}}, \Omega_{\text{op}}, \mathcal{M})$.
\IF{$C$ passes deterministic audit}
    \RETURN $C$
\ELSE
    \RETURN Refine constraint conflicts in $C$.
\ENDIF
\end{algorithmic}
\end{algorithm}

\subsubsection{Algorithm Agent}\label{subsec:micro-search-algorithm}

The algorithm agent conducts equal-budget strategy discovery in H1, but does not certify candidate validity. Its inputs are the research setting, remaining budget, candidate pool $\mathcal{P}_{t-1}$, and the algorithm view from shared memory. Its outputs are the next candidate strategies $\pi_t$ with identity controls $\pi_{\mathrm{ctrl},t}$. It first runs a model-eligibility probe to retain only comparable downstream models. It then applies local edits on the strategy graph, such as operator insertion, deletion, reordering, probability adjustment, and strength refinement. For each candidate, it constructs a matched control that differs only in the augmentation intervention. Based on validated historical evidence and the remaining budget, it decides which real training to run next. Candidate classes with repeated biological violations or execution failures are down-weighted or skipped, narrowing the search to an effective subset. Every candidate is trained with its identity control. A candidate advances only if it satisfies both predictive gain and biological feasibility. All failure, rejection, and repair outcomes are recorded in shared memory as negative evidence. The complete pseudocode is given in Algorithm~\ref{alg:micro-search}.

\begin{algorithm}[!htbp]
\caption{\small Multi-Agent Micro-Scale Searching}
\label{alg:micro-search}
\scriptsize
\begin{algorithmic}[1]
\REQUIRE Contract $C$, operator library $\Omega_{\text{op}}$, model set $\mathcal{M}$, budget $B$, pool size $K$.
\ENSURE Locked augmentation policy $\pi^{*}$ or abstention.
\STATE $\pi^{*} \leftarrow \emptyset$
\STATE Recall initial state $S_0$ from shared memory using $C$.
\STATE Probe model set $\mathcal{M}$ to select the best eligible model $m^{*}$.
\STATE Initialize candidate pool $\mathcal{P}_0$ from $\Omega_{\text{op}}$ and $S_0$.
\FOR{$t=1$ to $B$}
    \STATE Propose candidate $\pi_t$ from pool $\mathcal{P}_{t-1}$ and state $S_{t-1}$.
    \STATE Build matched identity control $\pi_{\text{ctrl},t}$ for $\pi_t$.
    \STATE Execute training and evaluation to obtain evidence $E_t$.
    \STATE Aggregate predictive gain $\Delta Y_t$ and biological constraints $\mathcal{C}_t$ from $E_t$.
    \STATE Make decision $d_t$ (accept, repair, or reject) based on $E_t$, $\Delta Y_t$, and $\mathcal{C}_t$.
    \IF{$d_t$ is accept}
        \STATE Confirm $\pi_t$ at full fidelity to obtain $\pi^{*}$.
        \STATE Publish $E_t$ and $\pi^{*}$ to shared memory.
    \ELSIF{$d_t$ is false-winner}
        \STATE Repair $\pi_t$ to $\pi_t'$ via paired counterfactual edits.
        \STATE Publish $E_t$ and repair outcome to shared memory.
    \ELSE
        \STATE Publish rejection evidence to shared memory.
    \ENDIF
    \STATE Update state $S_t$ from shared memory.
\ENDFOR
\IF{$\pi^{*}\neq\emptyset$ and $\pi^{*}$ passes locked test}
    \RETURN $\pi^{*}$
\ELSE
    \RETURN abstention
\ENDIF
\end{algorithmic}
\end{algorithm}

\subsubsection{Feedback Agent}\label{subsec:intervention}

The feedback agent provides auditable computation and independent adjudication. It receives the contract, evidence snapshot, metrics $\mathcal{Y}$, candidate lineage, and trial history $\mathcal{H}$. It compiles the constraint AST into executable predicates. It computes the paired gain $\Delta Y=Y(\pi)-Y(\pi_{\text{ctrl}})$ and the constraint vector $\mathcal{C}$. The vector covers Moran's $I$, Geary's $C$, marker-gene correlation, neighborhood consistency, and staining drift. Its outputs are the violation set, verdict, and stopping signal. A high-$\Delta Y$ candidate that violates constraints is flagged as a false winner. It is repaired by removing or weakening the suspected operator and rerunning under the same split, model, budget, and seed. The repair is accepted only if it passes constraints and keeps gain above the non-inferiority bound.

\subsubsection{Shared Memory Layer}\label{subsec:shared-memory}

The shared memory layer stores research settings, context signatures, candidate strategies, matched controls, raw execution artifacts, evidence, verdicts, and repair records. All outcomes are projected into role-restricted views. The modeling agent retrieves validated priors and constraint conflicts. The algorithm agent reads success rates, failure patterns, and transferable local edits. The feedback agent reads re-checkable execution records and historical verdicts. Let $w_t(a)$ denote the evidence weight of action $a$ under similar contexts, and let $v_t(a)$ denote its constraint-violation risk. The admissible action set of the next round is then
\begin{equation}
\footnotesize
\mathcal{A}(S_{t+1})=\left\{a\in\mathcal{A}(S_t): w_t(a)\geq\gamma,\;v_t(a)\leq\kappa\right\},
\label{eq:memory-pruning}
\end{equation}
where thresholds $\gamma$ and $\kappa$ are pre-declared in the research setting. Eq.~\eqref{eq:memory-pruning} captures the core of micro-scale searching, where memory serves not as a conclusion bank, but maps accumulated evidence to the next candidate scope. This avoids re-executing ineffective or high-risk edits.

\subsection{Evaluation Mechanism and Biological Consistency}\label{subsec:evaluation-mechanism}

The search kernel addresses how candidates are proposed, but not how their effectiveness is judged. In spatial transcriptomics, an augmentation may improve the validation metric while breaking the morphology-to-expression correspondence, yielding a false winner. Evaluation must therefore be an active component inside the search loop, not a terminal metric.

Predictive gain alone is insufficient for acceptance. From the same execution evidence, the feedback agent computes three consistency scores over image, spatial, and expression channels. The image-level check evaluates whether augmentation preserves tissue regions, pixel range, sharpness, and staining optical density. Let $X$ and $\tilde X_\pi$ be the original and augmented H\&E patches. Let $M(\cdot)$, $\operatorname{clip}(\cdot)$, $\operatorname{blur}(\cdot)$, and $\mu_{\mathrm{HED}}(\cdot)$ denote the tissue-region mask, pixel-clipping ratio, Laplacian-variance sharpness, and mean HED optical-density vector \cite{tellez2019quantifying}. Define $A=\operatorname{clip}(\tilde X_\pi)$, $B=\operatorname{blur}(\tilde X_\pi)-\operatorname{blur}(X)$, and $C=\mu_{\mathrm{HED}}(\tilde X_\pi)-\mu_{\mathrm{HED}}(X)$. Image consistency is then
\begin{equation}
\footnotesize
\begin{split}
q^{\mathrm{img}} = & w_{\mathrm{roi}} \mathrm{IoU}(M(X), M(\tilde X_\pi)) + w_{\mathrm{clip}}(1 - A/r_{\mathrm{clip}}) \\
& + w_{\mathrm{blur}}(1 - |B|/r_{\mathrm{blur}}) + w_{\mathrm{hed}}(1 - \|C\|_1/r_{\mathrm{hed}}).
\end{split}
\label{eq:image-consistency}
\end{equation}
The non-negative weights sum to one. The normalization scales are fixed on the training split and frozen during the search. The consistency scores are bounded within $[0,1]$.

Let $G$ and $\hat G$ be the ground-truth and predicted expression matrices on the validation split. All normalization scales are fixed on the validation set. Spatial consistency is characterized by the relative deviations of Moran's $I$ and Geary's $C$.
\begin{equation}
\footnotesize
q^{\mathrm{spa}}=1-\frac{1}{2}\left(\frac{|I(\hat G)-I(G)|}{r_I}+\frac{|C(\hat G)-C(G)|}{r_C}\right).
\label{eq:spatial-consistency}
\end{equation}
For a pre-declared set of marker-gene or gene-program pairs $\mathcal{P}$, expression-relation preservation is
\begin{equation}
\footnotesize
q^{\mathrm{exp}}=1-\frac{1}{|\mathcal{P}|}\sum_{(u,v)\in\mathcal{P}}\frac{|\rho_{\hat G}(u,v)-\rho_G(u,v)|}{r_\rho}.
\label{eq:expression-consistency}
\end{equation}
These three checks form the biological-consistency vector $\mathbf{q}_t=(q_t^{\mathrm{img}},q_t^{\mathrm{spa}},q_t^{\mathrm{exp}})$. The feedback agent compares $\mathbf{q}_t$ with the pre-declared threshold vector $\boldsymbol{\tau}=(\tau_{\mathrm{img}},\tau_{\mathrm{spa}},\tau_{\mathrm{exp}})$ and, together with the paired gain $\Delta Y_t$, issues an accept, repair, or reject decision.

\section{Experimental Setup and Results}\label{sec:experiments}
This section validates the three core propositions based on the following three research questions. 
\begin{itemize}
  \item RQ1: Can MMDA-BCST outperform state-of-the-art augmentations and other search strategies under a fixed budget?
  \item RQ2: Can biological-consistency feedback identify high-scoring yet biologically infeasible candidates as False Winners?
  \item RQ3: Can shared memory reduce repeated probing and improve search efficiency?
\end{itemize}

\subsection{Experimental Setup}\label{subsec:experiment-logic}

\subsubsection{Datasets and Evaluation Units}

We evaluated MMDA-BCST on three spatial transcriptomics cohorts. Table~\ref{tab:datasets} summarizes their statistics and evaluation protocols. For the Colon dataset, we performed case-level three-fold cross-validation. To estimate uncertainty, the validation case was divided into four spatial regions that were not treated as independent cases. The Gullet dataset contains a single sample (G1), so its three spatial regions support within-sample analyses only and contribute one sample-level value to the pooled summary. For HER2ST, we split the data by case into training, validation, and testing sets. Each HER2ST case was evaluated with three random seeds. The results were averaged within each case before calculating the between-case mean and standard deviation.

\begin{table}[t]
\centering
\caption{Summary of the evaluated datasets.}
\label{tab:datasets}
\footnotesize
\setlength{\tabcolsep}{3pt}
\begin{tabular}{lccc}
\toprule
\textbf{Property} & \textbf{Colon} & \textbf{Gullet} & \textbf{HER2ST} \\
\midrule
Cancer type & Colorectal & Esophageal & Breast \\
Source & Public & Non-public & Public \\
Cases & P1, P2, P5 & G1 & 8 cases (36 slices) \\
Observations & 5k per case & 3,000 & 13,620 \\
Genes & 18,085 & 6,175 & 11,871 \\
Patch Size & \(224\times224\) & \(224\times224\) & \(224\times224\) \\
Data Split & Case 3-fold & 3 regions & Case hold-out \\
Evaluation & Mean \(\pm\) std & Within-sample & 3 seeds/case \\
\bottomrule
\end{tabular}%
\end{table}

\subsubsection{Models, augmentation space, and search comparisons}

Downstream models are DeepSpot \cite{ratschlab2025deepspot}, HE2Gene \cite{he2020integrating}, NH2ST \cite{liu2025nh2st}, and TRIPLEX \cite{chen2024triplex}. Each model is evaluated without augmentation before the search. Public implementations are used. GigaPath encoders are frozen \cite{xu2024gigapath}, and ResNet backbones are ImageNet-pretrained. Fixed candidates include no augmentation, model-native pipeline, center-preserving geometry, CutMix \cite{yun2019cutmix}, RandAugment \cite{cubuk2020randaugment}, AutoAugment \cite{cubuk2019autoaugment}, AugMix \cite{hendrycks2020augmix}, TrivialAugment \cite{muller2021trivialaugment}, RandStainNA \cite{shen2022randstainna}, HED staining, MHOA\cite{liu2026mhoa}, MPA\cite{jang2026singlepatchbaseline}, GenST\cite{wood2026genst}, and SNR-ST-Mix\cite{yu2026snrstmix}. Search operators support insertion, deletion, reordering, and probability/strength adjustment. The maximum graph length is 3, and strengths are quantized into 31 levels. Occluding, mixing, and spatially altering operators are used for exploration only. Search methods include uniform random search, differential evolution, a single-agent variant, and full MMDA-BCST, all under the same candidate scope and budget.

\subsubsection{Evaluation protocol and implementation settings}

Evaluation follows paired comparison, staged screening, and locked testing. Each candidate is paired with a no-augmentation control under an identical data split, model, initial weights, data order, and seed. The search stage only performs pruning, while a candidate advances only if it achieves a paired GeneMean gain and passes biological checks. High-gain candidates violating constraints are repaired counterfactually, not accepted directly. For the reported strategies, the strategy graph, training parameters, and random settings are frozen. They are then retrained with controls on the reserved test set. GeneMean is the primary endpoint, the mean Pearson correlation on evaluation-eligible genes. Auxiliary metrics cover gene-set stability, spot-level behavior, and global error. Biological feasibility is enforced as an independent condition. Shared-memory ablation experiments compare no-history, success-only, and joint success/failure settings. Efficiency improvements are considered valid only if prediction and constraints do not degrade. Implementation details are provided in Table~\ref{tab:protocol-config}.

%\begin{table*}[t]
%\centering
%\caption[Summary of experimental configurations]{\centering \textbf{Summary of Evaluation and Implementation Configurations}}
%\label{tab:protocol-config}
%\scriptsize
%\setlength{\tabcolsep}{6pt}
%\renewcommand{\arraystretch}{1.25}
%\begin{tabular}{@{}L{3.0cm}L{14.2cm}@{}}
%\toprule
%\textbf{Configuration item} & \textbf{Setting} \\
%\midrule
%Primary endpoint & GeneMean (zeroed if prediction is constant but ground truth has variance). Report gene coverage and degradation ratio. \\
%Auxiliary endpoints & GeneMedian, GeneTop-$k$ ($k=50/100/250/1000$), RMSE, MAE. \\
%Biological checks & Image checks (tissue region, blur, clipping, HED drift). Expression checks (marker genes, co-expression programs). Spatial checks ($k=6$ adjacency graph). \\
%Search budget & At most 20 candidates per run with 4 alternates. Candidates and controls differ only in augmentation. \\
%Training stages & 20\%/50\%/full-data subsets for 10/25/50 epochs. Final candidate passes two full-training verifications. \\
%Acceptance condition & Paired GeneMean gain lower bound $>10^{-3}$ and all pre-declared biological checks pass. \\
%Random seeds & Augmentation runs use seeds 11, 23, and 47. Search comparisons use seed 11. \\
%Hardware and training & 2$\times$ RTX 4090 D. Early-stopping patience 10, batch size 64, best validation checkpoint. \\
%Statistical unit & Colon uses the patient as the independent unit. Gullet uses within-sample pairing and spatial-region bootstrapping. \\
%\bottomrule
%\end{tabular}
%\end{table*}

\begin{table}[t]
\centering
\caption{Summary of Evaluation and Implementation Configurations}
\label{tab:protocol-config}
\scriptsize
\setlength{\tabcolsep}{3pt}
\renewcommand{\arraystretch}{1.15}
\begin{tabular}{@{}p{2.3cm}p{6.2cm}@{}}
\toprule
\textbf{Item} & \textbf{Setting} \\
\midrule
Primary endpoint & GeneMean (zeroed if constant prediction vs.\ variable GT); report gene coverage and degradation ratio. \\
Auxiliary endpoints & GeneMedian, GeneTop-$k$ ($k=50/100/250/1000$), RMSE, MAE. \\
Biological checks & Image (tissue, blur, clipping, HED drift); expression (marker genes, co-expression); spatial ($k=6$ adjacency). \\
Search budget & $\le$20 candidates/run with 4 alternates; candidates and controls differ only in augmentation. \\
Training stages & 20\%/50\%/full data for 10/25/50 epochs; final candidate passes two full-training verifications. \\
Acceptance & Paired GeneMean gain lower bound $>10^{-3}$ and all pre-declared biological checks pass. \\
Random seeds & Augmentation: 11/23/47; search comparisons: 11. \\
Hardware \& training & 2$\times$ RTX 4090 D; early-stopping patience 10, batch 64, best validation checkpoint. \\
Statistical unit & HER2ST \& Colon: patient-level; Gullet: within-sample pairing and spatial-region bootstrap. \\
\bottomrule
\end{tabular}
\end{table}

\subsection{RQ1: Can MMDA-BCST outperform state-of-the-art augmentations and other search strategies under a fixed budget?}\label{subsec:framework-strategy-results}

This subsection addresses RQ1 by first comparing MMDA-BCST with state-of-the-art augmentation methods, and then evaluating MMDA-BCST against three other search strategies under an identical budget.

\subsubsection{Comparison with state-of-the-art augmentation methods}

\begin{figure}[!t]
    \centering
    \includegraphics[width=1\columnwidth]{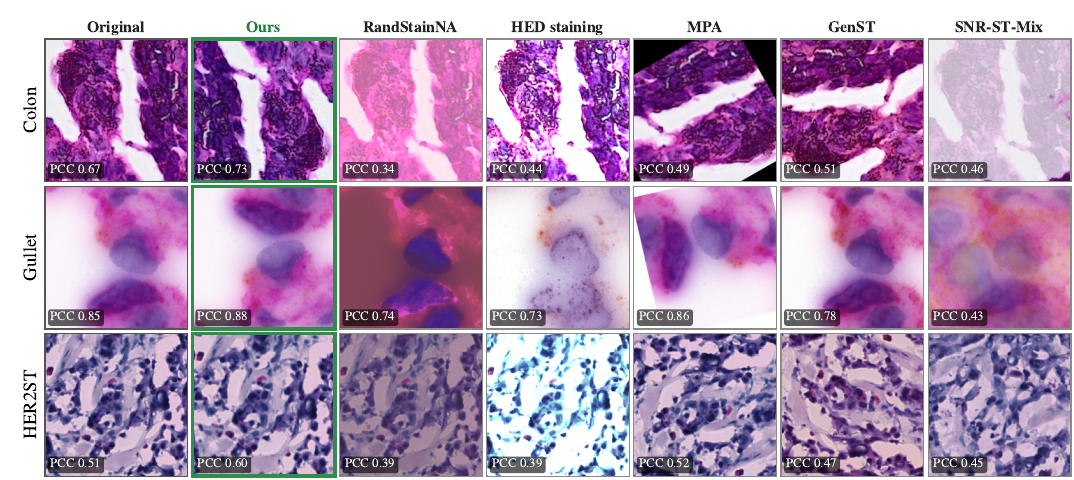}
\caption{The augmented patch samples based on MMDA-BCST and the compared state-of-the-art methods.}
    \label{fig:SOTA}
\end{figure}

Table~\ref{tab:augmentation-performance} presents a comparison between MMDA-BCST and the state-of-the-art methods under the same experimental protocol. The proposed MMDA-BCST achieves the best overall predictive performance, obtaining the highest GeneMean ($0.1860\pm0.0722$) and GeneTop50 ($0.5698\pm0.1566$). Although GenST is the strongest among all external augmentation strategies, MMDA-BCST achieves a relative improvement of 18.0\% in GeneMean and 17.3\% in GeneTop50 over it. Furthermore, relative to no augmentation, MMDA-BCST improves GeneMean and GeneTop50 by 23.2\% and 23.0\%, respectively. The large standard deviations suggest that augmentation performance depends on the sample and downstream model. These results provide fixed-strategy references for the subsequent search-method comparison under identical splits, models, and budgets. Beyond these quantitative metrics, qualitative visual examples further demonstrate the superiority of MMDA-BCST. As shown in Fig.~\ref{fig:SOTA}, several state-of-the-art augmentation methods produce patches with distorted tissue structures or degraded predictive performance. In contrast, MMDA-BCST consistently preserves tissue morphology and yields higher PCC scores across all three datasets.

\begin{table}[!h]
\caption{Exploratory Results of Fixed Augmentation Methods.}
\label{tab:augmentation-performance}
\centering
\scriptsize
\setlength{\tabcolsep}{2.5pt}
\renewcommand{\arraystretch}{1.08}
\resizebox{\columnwidth}{!}{%
\begin{tabular}{llcccc}
\toprule
Category & Augmentation strategy & GeneMean $\uparrow$ & GeneTop50 $\uparrow$ & \textcolor{red}{MSE $\downarrow$} & \textcolor{red}{MAE $\downarrow$} \\
\midrule
\textcolor{red}{Control} & No augmentation & \textcolor{red}{$0.1510\pm0.0678$} & \textcolor{red}{$0.4632\pm0.1625$} & \textcolor{red}{$1.7568\pm1.5102$} & \textcolor{red}{$0.1973\pm0.0770$} \\
\midrule
\multirow{7}{*}{\shortstack[l]{\textcolor{red}{General}\\\textcolor{red}{purpose}}} & Center-preserving geometry & \textcolor{red}{$0.1518\pm0.0703$} & \textcolor{red}{$0.4757\pm0.1406$} & \textcolor{red}{$1.8054\pm1.5930$} & \textcolor{red}{$0.1987\pm0.0779$} \\
 & AugMix\cite{hendrycks2020augmix} & \textcolor{red}{$0.1544\pm0.0678$} & \textcolor{red}{$0.5262\pm0.1474$} & \textcolor{red}{$1.7656\pm1.5431$} & \textcolor{red}{$0.1961\pm0.0776$} \\
 & TrivialAugment\cite{muller2021trivialaugment} & \textcolor{red}{$0.1486\pm0.0757$} & \textcolor{red}{$0.5137\pm0.1556$} & \textcolor{red}{$\mathbf{1.6797\pm1.5231}$} & \textcolor{red}{$0.1950\pm0.0783$} \\
 & RandAugment\cite{cubuk2020randaugment} & \textcolor{red}{$0.1408\pm0.0768$} & \textcolor{red}{$0.5025\pm0.1545$} & \textcolor{red}{$1.8819\pm1.5820$} & \textcolor{red}{$0.1967\pm0.0802$} \\
 & CutMix\cite{yun2019cutmix} & \textcolor{red}{$0.1398\pm0.0822$} & \textcolor{red}{$0.4861\pm0.1692$} & \textcolor{red}{$1.8065\pm1.5748$} & \textcolor{red}{$0.1985\pm0.0797$} \\
 & AutoAugment\cite{cubuk2019autoaugment} & \textcolor{red}{$0.1248\pm0.1059$} & \textcolor{red}{$0.4718\pm0.2004$} & \textcolor{red}{$1.8336\pm1.5248$} & \textcolor{red}{$0.2002\pm0.0832$} \\
 & MHOA\cite{liu2026mhoa} & \textcolor{red}{$0.1529\pm0.0712$} & \textcolor{red}{$0.5269\pm0.1512$} & \textcolor{red}{$1.6898\pm1.4965$} & \textcolor{red}{$0.1957\pm0.0770$} \\
\midrule
\multirow{5}{*}{\shortstack[l]{\textcolor{red}{ST-specific}}} & RandStainNA\cite{shen2022randstainna} & \textcolor{red}{$0.1536\pm0.0710$} & \textcolor{red}{$0.5332\pm0.1441$} & \textcolor{red}{$1.8167\pm1.5378$} & \textcolor{red}{$0.1983\pm0.0801$} \\
 & HED staining & \textcolor{red}{$0.1208\pm0.0585$} & \textcolor{red}{$0.4718\pm0.1158$} & \textcolor{red}{$1.9327\pm1.6154$} & \textcolor{red}{$0.1977\pm0.0795$} \\
 & MPA\cite{jang2026singlepatchbaseline} & \textcolor{red}{$0.1548\pm0.0735$} & \textcolor{red}{$0.5247\pm0.1614$} & \textcolor{red}{$1.7278\pm1.5698$} & \textcolor{red}{$0.1954\pm0.0775$} \\
 & GenST\cite{wood2026genst} & \textcolor{red}{$0.1576\pm0.0776$} & \textcolor{red}{$0.4859\pm0.1536$} & \textcolor{red}{$1.7342\pm1.5235$} & \textcolor{red}{$\mathbf{0.1935\pm0.0787}$} \\
 & SNR-ST-Mix\cite{yu2026snrstmix} & \textcolor{red}{$0.1548\pm0.0731$} & \textcolor{red}{$0.5271\pm0.1616$} & \textcolor{red}{$1.7114\pm1.5522$} & \textcolor{red}{$0.1970\pm0.0762$} \\
\midrule
\textbf{Ours} & \textbf{MMDA-BCST} & \textcolor{red}{$\mathbf{0.1860\pm0.0722}$} & \textcolor{red}{$\mathbf{0.5698\pm0.1566}$} & \textcolor{red}{$1.7663\pm1.5621$} & \textcolor{red}{$0.1937\pm0.0775$} \\
\bottomrule
\end{tabular}
}
\end{table}

\subsubsection{Equal-budget search and algorithm-agent ablation}

Four search methods share the same data splits, model set, budget cap, and seed 11. Each method runs once per Colon split, yielding 12 experiments in total. Full MMDA-BCST uses a role-separated algorithm agent. The three controls merge agent roles, sample uniformly, or follow fixed differential-evolution rules. Table~\ref{tab:e1-search-result} reports the raw GeneMean, its change from the matched no-augmentation control, and the number of candidate evaluations. All values are scaled by $10^{-3}$. The mean gain and relative gain are computed within each split and averaged over the two evaluated splits.

On these two splits, MMDA-BCST attains GeneMean values of 0.2154 and 0.0800, with paired gains of 0.0051 and 0.0066. The average gain is 0.0058, and the average relative gain is 5.69\%. The other methods achieve average gains of 0.0033 for evolutionary search, 0.0026 for random search, and 0.0022 for single-agent search. Although random and evolutionary search obtain higher raw GeneMean on the second split, their matched controls are also higher. MMDA-BCST yields the largest paired gain on both splits. On the third split, all four methods stop safely because no candidate preserves the image--expression pairing. The top-scoring candidate on the second split also triggers a biological inconsistency. RQ2 further demonstrates how such false winners are excluded by biological feedback.

\begin{table}[H]
\caption{Search-method comparison and algorithm-agent ablation under a common budget cap.}
\label{tab:e1-search-result}
\centering
\scriptsize
\setlength{\tabcolsep}{2.8pt}
\renewcommand{\arraystretch}{1.14}
\resizebox{\columnwidth}{!}{%
\begin{tabular}{lccccc}
\toprule
Search method & Evaluations $\downarrow$ & \shortstack{Split 1\\Raw / $\Delta$} & \shortstack{Split 2\\Raw / $\Delta$} & \shortstack{Mean $\Delta$\\$\uparrow$} & \shortstack{Mean relative gain\\$\uparrow$} \\
\midrule
Single agent & 21 & $211.7/{+}1.3$ & $76.4/{+}3.0$ & 2.2 & 2.35\% \\
Random search & 28 & $211.7/{+}1.3$ & $83.4/{+}3.9$ & 2.6 & 2.80\% \\
Non-agent evolutionary search & 24 & $213.0/{+}2.7$ & $83.4/{+}3.9$ & 3.3 & 3.12\% \\
\midrule
\textbf{MMDA-BCST (Ours)} & 24 & $\mathbf{215.4/{+}5.1}$ & $\mathbf{80.0/{+}6.6}$ & \textbf{5.8} & \textbf{5.69\%} \\
\bottomrule
\end{tabular}
}
\end{table}

\subsection{RQ2: Can biological-consistency feedback identify high-scoring yet biologically infeasible candidates as false winners?}\label{subsec:feedback-biological-results}

Higher predictive scores do not necessarily imply consistency with expression information. This subsection compares two selection rules on the same candidate set. One rule selects the highest predictive gain, and the other selects the highest gain among biologically compliant candidates. 

\begin{figure}[!t]
    \centering
    \includegraphics[width=1\columnwidth]{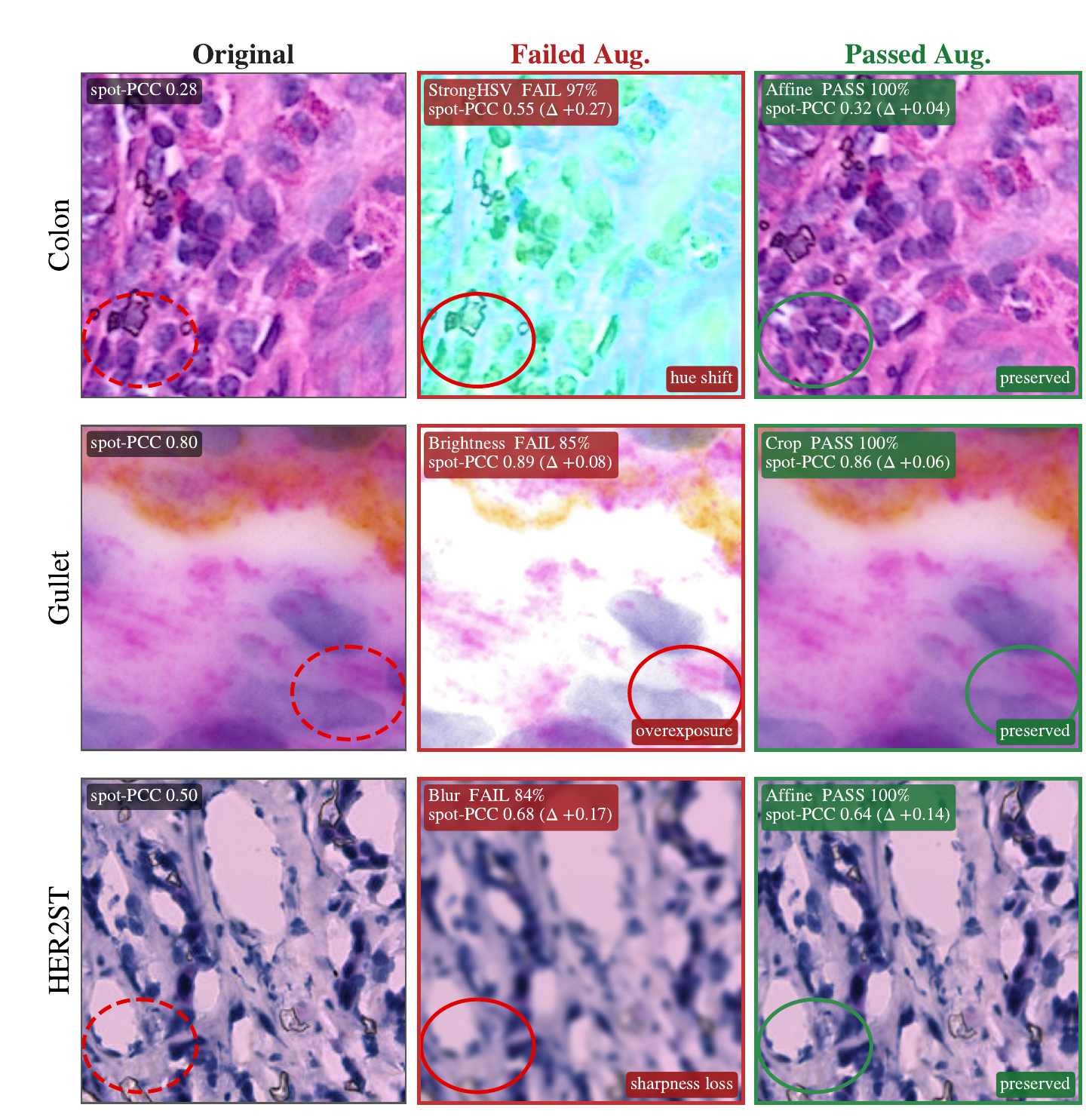}
\caption{\textbf{Representative patches illustrating biological-consistency feedback.} Original, failed, and passed augmentations are shown across three datasets. Spot-PCC and its change are reported for each patch. The morphology pass threshold is 80\%.}
    \label{fig:e3-visual-feedback}
\end{figure}

Fig.~\ref{fig:e3-visual-feedback} presents representative patches across three datasets. Failed augmentations like StrongHSV, Brightness, and Blur yield high gains but fail morphology checks on 84\% to 97\% of patches. They introduce severe hue shifts, overexposure, or sharpness loss. In contrast, affine and crop augmentations pass on all patches with smaller but reliable gains. These false winners are thus rejected by biological feedback. Overall, the two rules agree when metrics align, and the biologically compliant rule rejects the top scorer when they conflict.

The morphology pass rate requires each validation patch to simultaneously satisfy limits on tissue-region change, sharpness, pixel clipping, and HED drift. Data integrity is enforced as a separate mandatory condition. Predictive gain and morphological feasibility are kept distinct, so a high score cannot offset a morphological violation.

\begin{figure}[H]
    \centering
    \includegraphics[width=0.9\columnwidth]{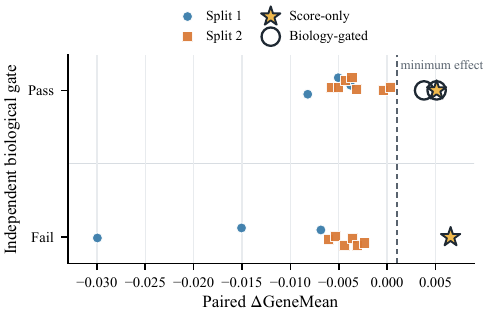}
    \caption{\textbf{Predictive gain and biological-check outcome for 20 candidate strategies.} The horizontal axis represents $\Delta$GeneMean against the matched control, and the vertical position indicates biological-check status with jitter. The dashed line marks $\Delta$GeneMean $=0.001$. The star and hollow circle denote prediction-only and biologically informed selections.}
    \label{fig:e3-prediction-biology}
\end{figure}

All 20 evaluated candidates are plotted in Fig.~\ref{fig:e3-prediction-biology} by their $\Delta$GeneMean and biological-check status. Both check-passing and check-failing candidates exhibit positive gains. High-scoring candidates that violate constraints are discarded, without globally reordering the remaining strategies. For the failed ColorJitter candidate, we tested four lower-perturbation variants by reducing strength, application probability, or both. All four variants yielded gains (0.00035, 0.00199, 0.00041, 0.00153) with confidence intervals crossing zero. The probability-reduced variant achieved the highest mean gain, yet its confidence interval still crossed zero. These results motivate the subsequent repair analysis and demonstrate that safe selection cannot rely solely on predictive gain.

\subsection{RQ3: Can Shared Memory Reduce Repeated Probing and Improve Search Efficiency?} \label{subsec:shared-memory-context-results}

Evaluating an augmentation strategy requires both predictive performance and biological consistency. Instead of transferring a globally optimal augmentation to all samples, evidence for each (context, model, augmentation, outcome) configuration is recorded in the shared memory. This process increases candidate priority under similar conditions and reduces blind-reuse risk under dissimilar ones. Panel A of Table~\ref{tab:context-summary} lists the fixed-augmentation map as contextual evidence. Panel B presents the paired ablation of experience reuse on computational overhead and predictive performance. The HER2ST entry extends this evidence to a multi-case setting and is interpreted separately from Colon and Gullet.

On Colon, the highest GeneMean for P1 and P2 is achieved by HE2Gene with the model-native pipeline, forming positive evidence for the same-domain context and raising the priority of similar candidates. On P5, the best combination changes to RandStainNA, which highlights case heterogeneity and limits cross-case extrapolation. Gullet's G1 is evaluated separately, where the best combination uses HE2Gene with center-preserving geometric augmentation. Because G1 is private, this evidence supports only within-sample transfer and similar-context initialization. It is not pooled with Colon. For HER2ST, HE2Gene with vertical-flip augmentation is recorded as the best contextual combination (GeneMean of 0.3222). This entry is used as dataset-specific evidence and is not combined with the other datasets.

\begin{table*}[!htbp]
\caption{\textbf{Contextual evidence in shared memory and its ablation results.} GeneMean values in Panel B are in units of $10^{-3}$. Colon and Gullet are interpreted separately.}
\label{tab:context-summary}
\centering
\scriptsize
\setlength{\tabcolsep}{3pt}
\renewcommand{\arraystretch}{1.05}

% ===== Panel A =====
\begin{tabular*}{\textwidth}{@{\extracolsep{\fill}}llccl@{}}
\toprule
\multicolumn{5}{l}{\textbf{Panel A: Contextual evidence from the fixed-augmentation map}} \\
\midrule
Data domain & Sample & GeneMean & GeneTop50 & Best combination per sample \\
\midrule
Colon & P1 & 0.0400 & 0.3071 & HE2Gene + model-native pipeline (0.0929) \\
Colon & P2 & 0.1462 & 0.6510 & HE2Gene + model-native pipeline (0.2508) \\
Colon & P5 & 0.0349 & 0.2738 & HE2Gene + RandStainNA (0.0772) \\
Gullet & G1 & 0.2497 & 0.7284 & HE2Gene + center-preserving geometry (0.3063) \\
HER2ST & 8 cases & 0.3222 & 0.4984 & HE2Gene + vertical flip (0.3222) \\
\bottomrule
\end{tabular*}

\vspace{1.5ex}

% ===== Panel B =====
\begin{tabular*}{\textwidth}{@{\extracolsep{\fill}}llcccc@{}}
\toprule
\multicolumn{6}{l}{\textbf{Panel B: Efficiency and performance ablation of shared memory}} \\
\midrule
Dataset & Method & Time (h) $\downarrow$ & Speedup $\uparrow$ & GeneMean $\uparrow$ (Raw / Change) & Memory util. $\uparrow$ (Dec. / Ent.) \\
\midrule
\multirow{2}{*}{Colon} & w/o Memory & $19.72 \pm 0.50$ & $1.00\times$ & $79.47 / 0.00$ & $0\% / 0\%$ \\
 & \textbf{w/ Memory (Ours)} & $\mathbf{1.95 \pm 0.06}$ & $\mathbf{10.09\times}$ & $\mathbf{81.09 \pm 1.42 / {+}1.62}$ & $\mathbf{92\% / 60\%}$ \\
\midrule
\multirow{2}{*}{Gullet} & w/o Memory & $5.20 \pm 0.07$ & $1.00\times$ & $6.68 \pm 7.52 / 0.00$ & $0\% / 0\%$ \\
 & \textbf{w/ Memory (Ours)} & $\mathbf{2.13}$ & $\mathbf{2.45\times}$ & $\mathbf{6.68 \pm 7.52 / 0.00}$ & $\mathbf{100\% / 80\%}$ \\
\midrule
\multirow{2}{*}{HER2ST} & w/o Memory & $20.62 \pm 10.36$ & $1.00\times$ & $123.9 \pm 175.4 / 0.00$ & $0\% / 0\%$ \\
 & \textbf{w/ Memory (Ours)} & $\mathbf{12.06 \pm 6.06}$ & $\mathbf{1.71\times}$ & $\mathbf{123.9 \pm 175.4 / 0.00}$ & $\mathbf{17\% / 25\%}$ \\
\bottomrule
\end{tabular*}
\end{table*}

On Colon, employing the shared memory reduces the average search time from 19.72 h to 1.95 h, a 10.09\(\times\) speedup. GeneMean rises from $79.47\times10^{-3}$ to $81.09\times10^{-3}$, a paired increase of $1.62\times10^{-3}$. The decision and entry utilization rates reach 92\% and 60\%, respectively. On Gullet, search time drops from 5.20 h to 2.13 h, a 2.45\(\times\) speedup, while mean GeneMean remains unchanged at $6.68\times10^{-3}$. The utilization rates reach 100\% and 80\%. For the five paired HER2ST cases, search time is reduced from $20.62\pm10.36$ h to $12.06\pm6.06$ h ($1.71\times$ speedup), while GeneMean remains unchanged at $123.9\pm175.4\times10^{-3}$. The utilization rates are 17\% and 25\%. Thus, the HER2ST subset exhibits reduced search time without a decrease in GeneMean, although its small number of paired cases limits the conclusion.

\section{Conclusion}\label{sec:conclusion}

This paper presented MMDA-BCST, a multi-agent framework driven by micro-scale searching for biologically consistent augmentation in spatial transcriptomics prediction. Augmentation-strategy discovery is formulated as a sequential decision problem under biological-consistency constraints. Three agents coordinated through a shared memory layer handle candidate generation, predictive evaluation, and biological verification. Positive and negative evidence accumulated across trials is reused to avoid redundant probing. Extensive experiments demonstrate that model-native pipelines and center-preserving geometric augmentation remain relatively stable, whereas strong perturbative augmentations provide no consistent benefit. Under an identical budget, the full multi-agent loop produces larger GeneMean gains than evolutionary, random, and single-agent search. A high-scoring yet biologically infeasible candidate is identified as a false winner, leading to the selection of a feasible alternative. Furthermore, incorporating the shared memory substantially reduces search time without degrading predictive performance. The gains arise not from increasing the number of agents, but from closing the multi-agent loop where predictive evidence and independent biological feedback jointly determine candidate advancement. This provides a practical way to incorporate biological constraints into augmentation search without manual recipe tuning.

Future work will examine generalization across additional organs, platforms, and staining conditions, with locked external validation on both array-based and imaging-based spatial transcriptomics data.

% References embedded for a single-file manuscript.
% Generated by IEEEtran.bst, version: 1.14 (2015/08/26)

\end{document}